\documentclass[runningheads]{Springer/llncs}
\usepackage{graphicx}
\usepackage{caption}
\usepackage{float} 
\usepackage{subcaption}
\usepackage{booktabs}
\usepackage{amsmath}
\usepackage{multirow}
\usepackage{hyperref}
\usepackage{bm}

\usepackage{xcolor}
\usepackage[misc]{ifsym} 

\begin{document}
\title{RT-KGD: Relation Transition Aware Knowledge-Grounded Dialogue Generation}
\titlerunning{RT-KGD: Relation Transition Aware KGD}
%
%
\author{Kexin Wang\inst{1}\thanks{The first two authors made equal contributions to this work.} \and
Zhixu Li\inst{2\star}  $^{(\textrm{\Letter})}$  \and
Jiaan Wang\inst{1} \and
Jianfeng Qu\inst{1}$^{(\textrm{\Letter})}$ \and
Ying He\inst{3} \and \\
An Liu\inst{1} \and
Lei Zhao\inst{1}
}
\authorrunning{K. Wang et al.}
\institute{
School of Computer Science and Technology, Soochow University, Suzhou, China \\
\email{\{kxwang1, jawang1\}@stu.suda.edu.cn},
\email{\{jfqu, anliu, zhaol\}@suda.edu.cn} \and
Shanghai Key Laboratory of Data Science, School of Computer Science, Fudan University, Shanghai, China \\
\email{zhixuli@fudan.edu.cn}\and
IFLYTEK Research, Suzhou, China \\
\email{yinghe@iflytek.com}}
\maketitle


\begin{abstract}
Grounding dialogue system with external knowledge is a promising way to improve the quality of responses. Most existing works adopt knowledge graphs (KGs) as the external resources, paying attention to the contribution of entities in the last utterance of the dialogue for context understanding and response generation. Nevertheless, the correlations between knowledge implied in the multi-turn context and the transition regularities between relations in KGs are under-explored. To this end, we propose a Relation Transition aware Knowledge-Grounded Dialogue Generation model (RT-KGD). Specifically, inspired by the latent logic of human conversation, our model integrates dialogue-level relation transition regularities with turn-level entity semantic information. In this manner, the interaction between knowledge is considered to produce abundant clues for predicting the appropriate knowledge and generating coherent responses. The experimental results on both automatic evaluation and manual evaluation indicate that our model outperforms state-of-the-art baselines.

\end{abstract}
\section{Introduction}

Knowledge-Grounded Dialogue Generation (KGD) aims at generating an informative response based on both dialogue context and external knowledge~\cite{Ghazvininejad2018AKN,Kim2020SequentialLK}.
Current works typically utilize structured knowledge graphs (KGs)~\cite{Zhou2018CommonsenseKA,Moon2019OpenDialKGEC,Zhang2020GroundedCG} or unstructured texts~\cite{Zhao2020KnowledgeGroundedDG,Kim2020SequentialLK} as knowledge resources.
Incorporating external knowledge related to the dialogue context has proven to alleviate generating meaningless and bland responses caused by traditional generative models, such as ``\textit{I don't know}'' and ``\textit{You are right}''~\cite{Li2016ADO}.

\begin{figure}[t]
\centering
\includegraphics[width=\textwidth]{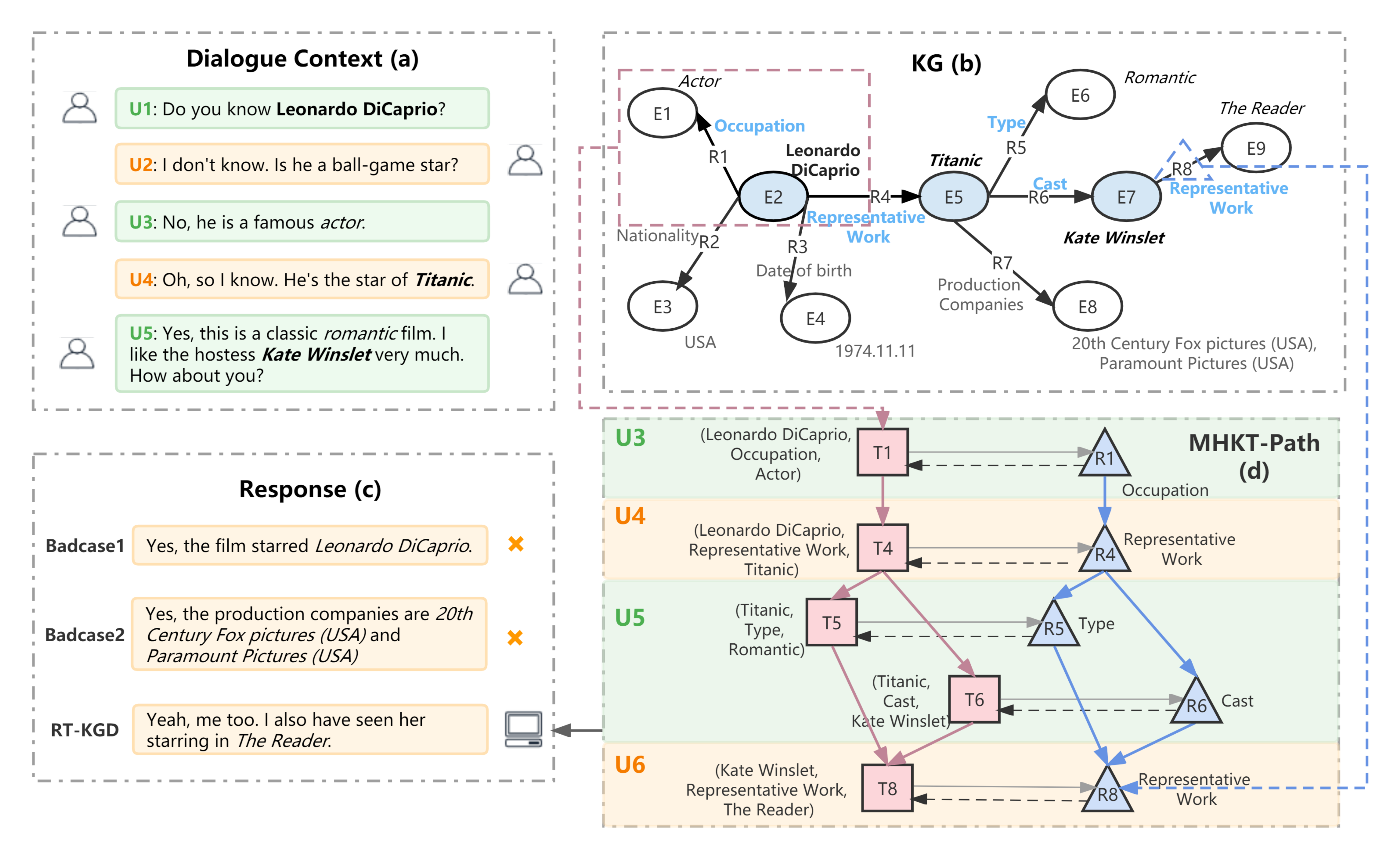}
\caption{An illustrative example from KdConv~\cite{Zhou2020KdConvAC}. Based on the dialogue context (a) and the related KG (b), KGD is required to generate a response (c) guided by the \texttt{MHKT-Path} (d). The \textbf{bold} denotes the core entities in the dialogue, and the \textit{Italic} denotes related knowledge values involved in the dialogue.
}
\label{fig:eg}
\end{figure}

The existing works mainly focus on two aspects in KGD task: knowledge-enhanced context understanding~\cite{Wu2020DiverseAI,Bai2021LearningTC} and knowledge-fused response generation~\cite{Lin2020GeneratingIC,Liang2021InfusingMK}.
%
Traditional efforts~\cite{Ghazvininejad2018AKN,Zhou2020KdConvAC,Bai2021LearningTC} simply treat the relevant external knowledge as the textual complementary to the dialogue context for both context understanding and response generation, neglecting considerable structural information in KGs.
%
Some recent works~\cite{Moon2019OpenDialKGEC,Zhang2020GroundedCG,Jung2020AttnIOKG} realize that the correlation between entities plays an important role in continuing dialogue, thus propose to excavate the valuable structural information between entities in the knowledge graph to predict the entities that might appear in the next response. The predicted entities are further used to guide the response generation.
For example, DialKG Walker~\cite{Moon2019OpenDialKGEC} treats the entities mentioned in the last utterance as the starting nodes and further retrieves relevant entities from KG within two hops.
DuConv~\cite{Wu2019ProactiveHC} pre-defines a topic goal including two entities for each dialogue, which guides the model to start with the first entity and gradually transition to the second one.

Despite their great contributions, there are two main drawbacks: on the one hand, the {\bf entity-guided KGD} methods~\cite{Moon2019OpenDialKGEC,Zhang2020GroundedCG} consider the entities in the dialogue as the only guidance knowledge for context understanding and response generation, which neglects the importance of {\bf relations} between entities in the KG.
However, the regularity behind human conversation can be summarized as a sequence of topics, where each topic may correspond to a relation between entities rather than a single entity in the KG.
%
%
%
On the other hand, the existing KGD methods~\cite{Moon2019OpenDialKGEC,Jung2020AttnIOKG} only care about the information in the last dialogue turn for predicting the subsequent knowledge, which is insufficient to learn how human transfer topics across a multi-turn dialogue. 
%
Taking Fig.~\ref{fig:eg} as an example, both badcase 1 and badcase 2 are flawed generated results based on the dialogue context.
Badcase 1 demonstrates that the generated response might be redundant and incoherent without modeling multiple turns of knowledge, while badcase 2 reveals an abrupt transition in the topic since the latent relation transition path throughout the dialogue is ignored.

In this paper, we propose a novel KGD model: Relation Transition aware Knowledge-Grounded Dialogue Generation (\texttt{RT-KGD}), which models the knowledge transition across multi-turn dialogue by integrating dialogue-level relation transition regularities with turn-level entity semantic information.
Specifically, we obtain all the relations and entities contained in the multi-turn dialogue context to construct a so-called Multi-turn Heterogeneous Knowledge Transition Path (\texttt{MHKT-Path}), which can be viewed as a subgraph of the external KG integrated with the sequential information of relations and entities in the multi-turn dialogue.
Based on the constructed \texttt{MHKT-Path}, a knowledge prediction module is proposed to retrieve the triplets that might appear in the subsequent response from the external KG, and they are further fused for triplet prediction.
Finally, the subsequent response is generated conditioned on both dialogue context and the predicted triplet.
As the example shown in Fig.~\ref{fig:eg}, the \texttt{MHKT-Path} grasps the latent conversation regularity of human beings, and the generated response based on the proposed \texttt{RT-KGD} is informative and coherent with the dialogue context.


%
The main contributions of this paper are concluded as follows:

\begin{itemize}
   \item To the best of our knowledge, we are the first to incorporate the relation transition across multi-turn dialogue into the KGD task. In this manner, the regularity behind human conversation can be portrayed by integrating relation transition paths and entity semantic information.

    \item We propose to build a Multi-turn Heterogeneous Knowledge Transition Path (MHKT-Path) for each dialogue, which integrates the structure information of external KG and the sequential information of knowledge with the multi-turn dialogue. Based on MHKT-Path, our model then retrieves appropriate knowledge from the KG to guide the next response generation.

    \item The experimental results on a multi-domain knowledge-driven dialogue dataset (i.e., KdConv~\cite{Zhou2020KdConvAC}) indicate that our model outperforms strong baseline models in both automatic and manual evaluation.

\end{itemize}


\section{Related Work}
\label{sec:2}

According to whether to introduce knowledge, we categorize previous dialogue generation works into \textit{Vanilla Dialogue Generation} and \textit{Knowledge-grounded Dialogue Generation}.

\noindent \textbf{Vanilla Dialogue Generation.} 
Early dialogue systems typically employ Sequ\\ence-to-Sequence (Seq2Seq) models to generate responses~\cite{Sordoni2015ANN,Serban2016BuildingED,Xing2018HierarchicalRA}, which is further improved with advanced context encoders~\cite{Serban2016BuildingED,Xing2018HierarchicalRA} or more efficient response generation methods~\cite{Zhang2018LearningTC,Bai2021LearningTC,zheng-etal-2021-enhancing-visual}. 
Recently, pre-trained generative models with the backbone of Transformer~\cite{Vaswani2017AttentionIA}, such as GPT-2~\cite{Radford2019LanguageMA} and BART~\cite{Lewis2020BARTDS}, achieve promising performance in many text generation tasks.
There is increasing work focusing on designing Transformer-based pre-trained dialogue models. Among them,
Blender~\cite{Roller2021RecipesFB} enhances Transformer architecture and show their superiority in dialogue generation.
DialoGPT~\cite{Zhang2020DIALOGPTL} extends GPT-2~\cite{Radford2019LanguageMA} for response generation.
Besides, PLATO~\cite{Bao2020PLATOPD} pre-trains unified language models for both bi-directional encoding and uni-directional decoding.
Nevertheless, they can only implicitly learn dialogue strategies and commonsense knowledge from dialogue corpora, resulting in limited transferability to other dialogue scenes.

\vspace{0.5ex} 
\noindent \textbf{Knowledge-Grounded Dialogue Generation.} 
A promising way to generate meaningful and informative responses is to utilize external knowledge to guide the models.
Generally, the external knowledge comes from textual corpora~\cite{Kim2020SequentialLK}, commonsense knowledge graphs~\cite{Zhou2018CommonsenseKA,Zhang2020GroundedCG,Wu2020DiverseAI}, and domain knowledge graphs~\cite{Wu2019ProactiveHC,Zhou2020KdConvAC}.
To utilize the knowledge, ~\cite{Vougiouklis2016ANN,Ghazvininejad2018AKN} adapt the memory network~\cite{Sukhbaatar2015EndToEndMN} to store the relevant knowledge and then generate responses conditioned on both dialogue context and stored knowledge.
Besides, ~\cite{Lian2019LearningTS,Wu2020DiverseAI} employ the posterior distribution of knowledge to guide its prior distribution, leading to accurate knowledge selection and high-quality generated responses.
Furthermore, some work~\cite{Lin2020GeneratingIC,Wu2020DiverseAI,Liang2021InfusingMK} leverages copy mechanism to copy words from knowledge sources directly and generate more informative responses.
Although great progress has been made, the structural information of KG is neglected, which might lead to suboptimal responses.

To effectively excavate the structural information, some researchers attempt to utilize graph neural networks on KG to obtain its structure-aware representation that is further incorporated into dialogue generation~\cite{Zhou2018CommonsenseKA,Moon2019OpenDialKGEC,Zhang2020GroundedCG}.
AttnIO~\cite{Jung2020AttnIOKG} leverages bi-direction attention flows to propagate messages from the entities appearing in the last utterance to their neighbor entities in KG.
%
ConceptFlow~\cite{Zhang2020GroundedCG} applies a graph attention mechanism to attend to appropriate concepts conditioning on dialogue context for responses generation, where the concepts are extracted from ConceptNet~\cite{Speer2017ConceptNet5A}, a large-scale commonsense knowledge graph.
Unlike previous research, our \texttt{RT-KGD} (1) refines the dialogue-level knowledge transition from different granularity; (2) incorporates the related knowledge based on the whole dialogue context rather than only the last utterance.

\section{Methodology}
\label{sec:3}
\begin{figure}[t]
\centering
\includegraphics[width=\textwidth]{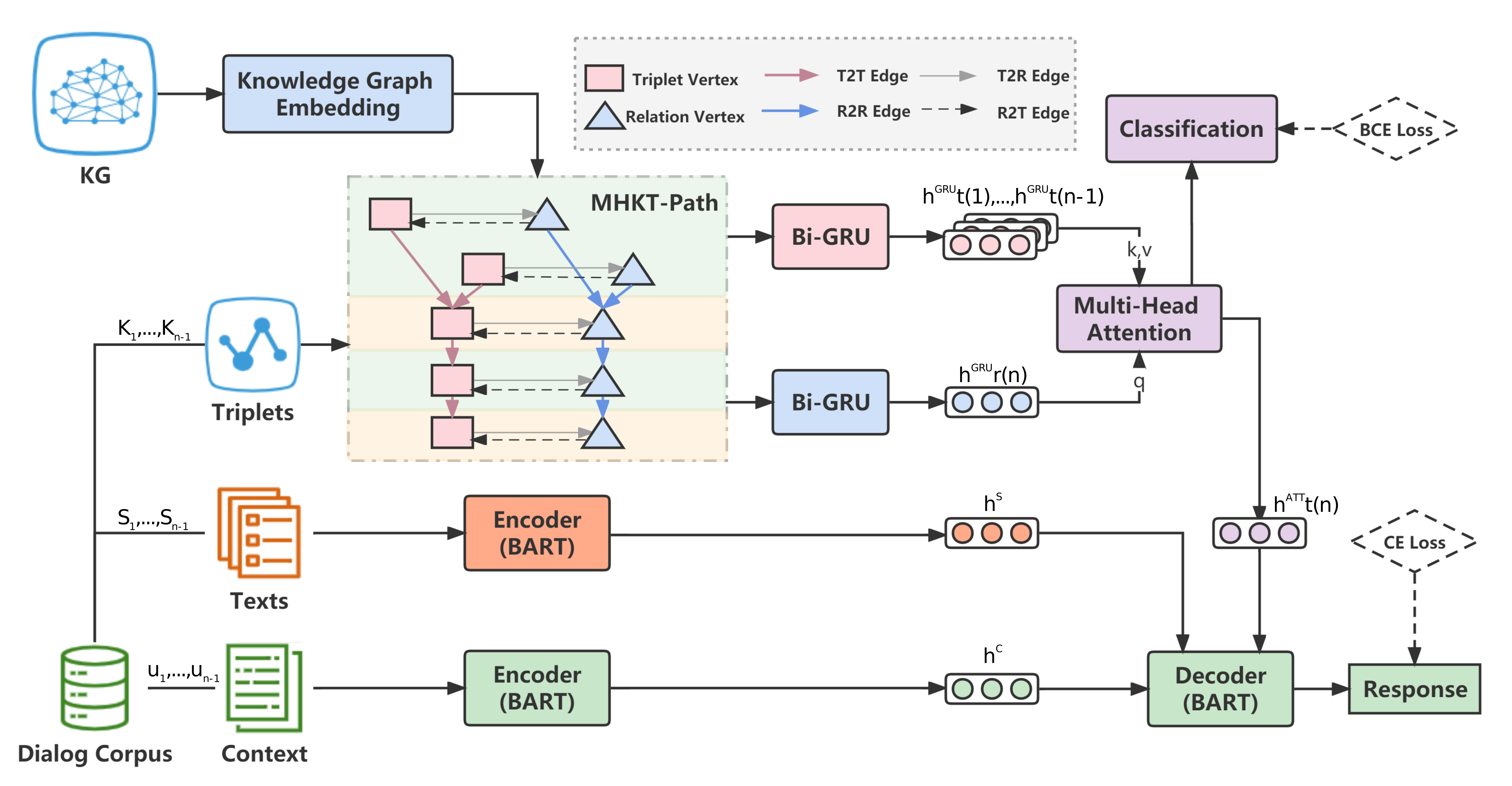}
\caption{The architecture of the proposed \texttt{RT-KGD} model.} \label{fig:model}
\end{figure}

In this section, we formally define the knowledge-ground dialogue generation task (Sec.~\ref{sec:3.1}) and then elaborate on four principal components of our \texttt{RT-KGD} model.
As illustrated in Fig.~\ref{fig:model}, our model first constructs the multi-turn heterogeneous knowledge transition path (\texttt{MHKT-Path}) for the given dialogue context (Sec.~\ref{sec:3.2}) and then encodes the \texttt{MHKT-Path} by a knowledge encoder (Sec.~\ref{sec:3.3}). Next, the predicted triplet from a knowledge prediction (Sec.~\ref{sec:3.4}) is finally incorporated into the subsequent response, which is generated by a knowledge-enhanced encoder-decoder (Sec.~\ref{sec:3.5}).

\subsection{Task Formulation}
\label{sec:3.1}
Given a dialogue context $C=\{u_1, u_2,\cdots,u_{n-1}\}$, where $u_{i}$ represents the $i$-th utterance.
Each $u_{i}$ corresponds to a knowledge triplet set $K_{i}=\{(h_{i_1},r_{i_1},t_{i_1}), (h_{i_2},\\r_{i_2},t_{i_2}),\cdots,(h_{i_{|K_{i}|}},r_{i_{|K_{i}|}},t_{i_{|K_{i}|}})\}$ $(|K_{i}|\geq0)$, 
where $(h,r,t)$ means that head entity $h$ and tail entity $t$ have a relation $r$, 
and a descriptive text set $S_{i}=\{s_{i_1},s_{i_2},\cdots,s_{i_{|S_{i}|}}\}$ $(|S_{i}|\geq0)$.
All knowledge triplets and descriptive texts are from domain knowledge graph $\mathcal{G}$ and corpus $\mathcal{O}$.
The goal of knowledge-grounded dialogue systems is to generate a proper response $u_n$ based on the dialogue context $C$, knowledge graph $\mathcal{G}$, and knowledge corpus $\mathcal{O}$.

\subsection{Multi-turn Heterogeneous Knowledge Transition Path}
\label{sec:3.2}

To integrate dialogue-level relation transition regularities with turn-level entity semantic information,
we utilize the knowledge triples associated with the given dialogue context, i.e., $K=K_{1} \cup K_{2} \cup \cdots \cup K_{n-1}$, to construct the multi-turn heterogeneous knowledge transition path, which is called \texttt{MHKT-path}. As shown in Fig.~\ref{fig:model}, \texttt{MHKT-path} contains two types of vertices, i.e., triplet vertices and relation vertices.
In detail, each triplet vertex represents a knowledge triplet belonging to $K$, and corresponds with a relation vertex which is extracted from it.
%
There are four types of edges in \texttt{MHKT-Path}: (1) the \textit{triplet-to-triplet} edge links the triplet vertices associated in one utterance with others in the neighbor utterances; (2) the paired \textit{triplet-to-relation} and (3) \textit{relation-to-triplet} edges denote the bi-directional interaction between triplet vertices and their corresponding relation vertices; (4) the \textit{relation-to-relation} edge links relation vertices with each other only if their corresponding triplet vertices are connected.
In this manner, the knowledge transition of both turn-level triplets and dialogue-level relations is integrated into the \texttt{MHKT-Path}.

\subsection{Knowledge Encoder}
\label{sec:3.3}
The knowledge encoder learns the representation of the vertices in \texttt{MHKT-Path}. Specifically, it contains vertex initializer and graph layers to initialize and update the vertex representations.

\subsubsection{Vertex Initializer} 
Instead of directly using the average word embeddings of the flat texts in entities and relations, we employ a KG embedding algorithm (i.e., TransR~\cite{Lin2015LearningEA}) to initialize the representation of vertices in our \texttt{MHKT-Path}\footnote{We also attempt to encode entities and relations based on word embedding, as suggested by~\cite{10.1145/3488560.3498405,10.1007/978-3-031-00129-1_6}, the results underperform that of using TransR.}:
\begin{equation}
    h^0_{e_{i}}=\textrm{TransR}(e_{i})
\end{equation}
where $e_{i}\in \mathcal{K}$ denotes a KG element (e.g., entity or relation), $h^0_{e_{i}}$ means the initialized representation of $e_{i}$. \textrm{TransR(·)} represents the TransR KG embedding function, learned by projecting entities from entity space to different relation spaces and building translations between the projected entities. 
In this way, the learned representation of KG elements in $\mathcal{K}$ contain the global KG structural information due to their interaction in KG~\cite{Lin2015LearningEA,Bordes2013TranslatingEF,10.1007/978-3-031-00129-1_10}.

For relation vertex in \texttt{MHKT-Path}, we directly use $h^0_{e_{i}}$ as its initial representation. 
For triplet vertex $(h_{i},r_{i},t_{i})$, we calculate its representation as:
\begin{equation}
\textrm{TransR}(h_{i}) \oplus \textrm{TransR}(r_{i}) \oplus \textrm{TransR}(t_{i})
\end{equation}
where $\oplus$ denotes concatenation.

\subsubsection{Graph Layers} 
Graph layers are used to update the vertex representations with the local structural information in the established \texttt{MHKT-Path}. Here, we employ the Heterogeneous Graph Transformer (HGT)~\cite{Hu2020HeterogeneousGT} as the graph layers since it is aware of different types of vertices and edges. Given the \texttt{MHKT-Path}, the representation of each vertex $v_{i}$ is updated by aggregating its neighbor information:

\begin{equation}
    HGT(h^{\ell}_{v_{i}})=\mathop{\mathbf{Aggregate}}\limits_{\forall v_{src}\in N(v_{i})}\bigg( \mathbf{Attention}(v^{\ell-1}_{i},v^{\ell-1}_{src})\cdot \mathbf{Message}(v^{\ell-1}_{src}) \bigg)
\end{equation}
where $N(v_{i})$ is the neighbor vertices set of $v_{i}$, the $\textbf{Aggregate(·)}$, $\textbf{Attention(·)}$, and $\textbf{Message(·)}$ are three basic operators in HGT:

\begin{itemize}
\item $\textbf{Attention(·)}$ calculates the mutual attention of each vertex pair, where each type of vertex and edge has a unique linear projection.

\item $\textbf{Message(·)}$ transfers information from different types of neighbor vertices of each vertex $v_i$.

\item $\textbf{Aggregate(·)}$ integrates messages from neighbor vertices with attention weights to the core vertex $v_i$.
\end{itemize}

Finally, for vertex $v_{i}$, we concatenate the final node representation and the corresponding initial node representation with a simple linear projection: 
\begin{equation}
    h_{v_{i}}^{HGT}=W([h^{0}_{v_{i}} \oplus h^{L}_{v_{i}}])
\end{equation}
where $W$ is trainable parameters, L is the number of layers of HGT.

\subsection{Knowledge Predictor}
\label{sec:3.4}

After obtaining the final representations of both triplet and relation vertices in \texttt{MHKT-Path}, the knowledge predictor is used to predict the knowledge which might be implied in the response.
There are three parts to knowledge prediction, i.e., relation prediction, relation-aware triplet prediction, and multi-label triplet classification.
Since the knowledge encoder aggregates only local neighborhood information, 
we further employ the bi-directional gated recurrent unit (Bi-GRU)~\cite{Cho2014LearningPR} to enrich the sequential representations of relations and triplets.

In detail, we first treat the average vertices representation in dialogue order as the input of Bi-GRU. 
Suppose there are $m$ relation vertices and $m$ triplet vertices in turn $i$. 
The relation vertices in turn $i$ are denoted as
$\{r_{i,j}\}_{j=1}^m$,
whose average representation is shown as follows:
\begin{equation}
    R^0_i=\mathop{Mean}(h^{HGT}_{r_{i,1}},\cdots,h^{HGT}_{r_{i,m}})
\end{equation}

Similarly, the triplet vertices in turn $i$ are denoted as $\{t_{i,j}\}_{j=1}^m\subset \mathcal{K}$, whose average representation is:
\begin{equation}
    T^0_i=\mathop{Mean}(h^{HGT}_{t_{i,1}},\cdots,h^{HGT}_{t_{i,m}})
\end{equation}

\subsubsection{Relation Prediction}
The relation prediction part is to obtain the $n$-th relation hidden state $h^{GRU}_r(n)$ based on the previous $n-1$ turns relation representation.
At step $t$ of relation prediction, 
Bi-GRU generates the $t$-th relation hidden state as follows:
\begin{equation}
\begin{aligned}
     h^{GRU}_r(t) &=[h^{fw}_r(t);h^{bw}_r(t)] \\
     &=[\mathop{GRU}\limits^\rightarrow(R^0_t,h^{fw}_r(t-1));\mathop{GRU}\limits^\leftarrow(R^0_t,h^{bw}_r(t-1))]
     \label{eq:GRU}
\end{aligned}
\end{equation}

\subsubsection{Relation Transition Aware Triplet Prediction}
Different from the relation, we utilize Bi-GRU to obtain $n-1$ triplet hidden states $h^{GRU}_t(1),\cdots,h^{GRU}_t(n-1)$ based on the input $T^0={T^0_1,\cdots,T^0_{n-1}}$. 
For the $i$-th triplet, its hidden state is calculated as follows:
%
\begin{equation}
\begin{aligned}
     h^{GRU}_t(i) &=[h^{fw}_t(i);h^{bw}_t(i)] \\
     &=[\mathop{GRU}\limits^\rightarrow(T^0_i,h^{fw}_T(i-1));\mathop{GRU}\limits^\leftarrow(T^0_i,h^{bw}_t(i-1))]
\end{aligned}
\end{equation}

After obtaining the predicted $n$-th relation hidden state and $n-1$ triplet hidden states, we employ multi-head attention~\cite{Vaswani2017AttentionIA} to jointly attend to the information from both dialogue level and turn level.
Thus the predicted triplet representation $h^{ATT}_{t_n}$ is calculated as follows:
\begin{equation}
\begin{aligned}
    \alpha_{i}&=\mathop{softmax_i}\bigg({h^{GRU}_{r_n}}^Th^{GRU}_{t_i}\bigg) \\
    h^{ATT}_{t_n}&=\mathop{\Vert} \limits_{d=1}^{D}\sum_{i=1}^{n-1}\alpha_i^{d}h^{GRU}_{t_i}
\end{aligned}
\end{equation}
where $D$ denotes the number of attention heads.

\subsubsection{Multi-label Triplet Classification} 
Since there might be multiple knowledge in the next response, the multi-label classification is adapted to map the predicted triplet representation to a label vector, where the number of labels is the total number of triplets in the knowledge graph $\mathcal{G}$.

Formally, let label $l=W_l(h^{ATT}_{t_n})\in \mathcal{R}^{|\mathcal{K}|}$, where $W_l$ is a trainable parameter and $|\mathcal{K}|$ is the total triplet size.
The target label is denoted as $y\in \{0,1\}^{|\mathcal{K}|}$.
%
%
Then we adapt the binary cross-entropy (BCE) loss to supervise the classification of triplets:
%
\begin{equation}
    \mathop{L_{BCE}}=-\frac{1}{\mathcal{K}}\sum_{i=1}^{\mathcal{K}}\bigg[y_ilog(\sigma(l_i))+(1-y_i)log(1-\sigma(l_i))\bigg]
\end{equation}
where $\sigma(\cdot)$ is sigmoid function.

\subsection{Knowledge-Enhanced Encoder-Decoder}
\label{sec:3.5}
We employ pre-trained BART~\cite{Lewis2020BARTDS} as the backbone of our KGD model, which aims to generate the final response based on dialogue context $C$, predicted triplet representation $K$ and corresponding descriptive texts $S$.
The input dialogue context is formed as ``\texttt{[CLS]${u_1}$[SEP]$u_{2}$[SEP]$\cdots$[SEP]$u_{n-1}$[SEP]}'', where \texttt{[CLS]} and \texttt{[SEP]} are two special tokens to indicate the utterance boundaries.
Then, the input is automatically tokenized by the BART's tokenizer, followed by a stack of BART encoder layers. Next, the context-aware representation of each token is obtained from the output of the last encoder layer of BART:
\begin{equation}
h^C_1,\cdots,h^C_{|C_{inp}|} = BART_{enc}(C)
\end{equation}
where $|C_{inp}|$ indicates the number of tokens in the input sequence, $BART_{enc}(\cdot)$ denotes the BART encoder, and $h^{C}_{i}$ is the context-aware representation of the $i$-th token in the sequence.

Similarly, for the descriptive text set $S=\{S_1,S_2,\cdots,S_{n-1}\}$ corresponding to the context $C$, each $S_i$ is encoded by the BART encoder, where the input is formed as ``\texttt{[CLS]$S_{i_1}$[SEP]$S_{i_2}$[SEP]$\cdots$[SEP]$S_{i_{|S_i|}}$}''.
We take the context-aware final representation of \texttt{[CLS]} as the sentence representation, and the encoded sentence embedding of the $i$-th turn is obtained as follows:
\begin{equation}
h^S_{i} = BART_{enc}(S_{i})
\end{equation}

Finally, the response is generated by the BART decoder, conditioning on the BART-encoded dialogue context $h^C_1,\cdots,h^C_{|C_{inp}|}$, descriptive sentences $h^S_{1},h^S_{2},\cdots\\,h^S_{|n-1|}$ and predicted triplets $h^{ATT}_{t_n}$: 
\begin{equation}
    G=BART_{dec}([h^C_1;h^C_2\cdots;h^C_{|C_{inp}|};h^S_{1};h^S_{2};\cdots;h^S_{|n-1|};h^{ATT}_{t_n};])
\end{equation}
where $G$ is the representation of generated response, $BART_{dec}(\cdot)$ denotes the BART decoder, $;$ denotes the token boundaries.

\subsubsection{Cross Entropy Loss} 
We guide the decoder with the ground-truth response $Y=u_n$ by computing the Cross-Entropy Loss:
\begin{equation}
    \mathop{L_{CE}}=-\frac{1}{|Y|}\sum_{t=1}^{|Y|}log(P(G_t=Y_t))
\end{equation}
where $G_t$ denotes the generated token at the decoding time step $t$, while $Y_t$ is the $t$-th token of the ground-truth response.
In summary, the final loss is defined by:
\begin{equation}
    \mathop{L_{total}}=\mathop{L_{CE}}+\lambda \cdot \mathop{L_{BCE}}
    \label{eq:loss}
\end{equation}
where $\lambda$ denotes the coefficients of the BCE loss.

\section{Experiments}
\label{sec:4}
\subsection{Dataset}
To verify our model, two requirements should be met in the datasets: (1) each utterance is annotated with related knowledge triples, and (2) containing abundant utterances in each dialogue.
Therefore, we conduct our experiments on KdConv\cite{Zhou2020KdConvAC}, a Chinese multi-domain knowledge-driven dialogue dataset, which contains 4.5K dialogues together with 86K utterances from three domains (i.e., film, music, and travel).
In KdConv, each dialogue contains 19.0 turns as well as 10.1 triplets on average.
For domain-specific knowledge, both structured triplets and unstructured texts are provided.
Specifically, the film, music, and travel domain knowledge contain 89K, 56K, and 10K triplets, together with 7.3K, 4.1K, and 1.1K descriptive sentences, respectively.
%

\subsection{Settings}

\subsubsection{Baselines:}
We adopt both vanilla and knowledge-grounded (indicating by ``+know'') dialogue generation models as our baselines:

\begin{itemize}

    \item \textbf{Seq2Seq}~\cite{Sutskever2014SequenceTS}: An encoder-decoder model augmented with attention mechanism~\cite{Bahdanau2015NeuralMT}.
    
    \item \textbf{Seq2Seq+know}~\cite{Zhou2020KdConvAC} fuses the last hidden state of the encoder with the knowledge vector via the attention mechanism and feeds both of them into the Seq2Seq decoder.
    
    \item \textbf{HRED}~\cite{Serban2016BuildingED}: A hierarchical recurrent encoder-decoder model which models utterances and context separately with different RNNs. 
    
    \item \textbf{HRED+know}~\cite{Zhou2020KdConvAC} fuses the context vector with the knowledge vector and treats the fused vector as the initial state of the HRED decoder.
    
    \item \textbf{BART}~\cite{Lewis2020BARTDS}: A pre-trained Transformer-based encoder-decoder model which achieves state-of-the-art performance on various text generation tasks.
    
    \item \textbf{BART+know} incorporates both knowledge entities and relations represented by the average word embeddings of the corresponding flat texts.
    
    \item \textbf{BART+know(TransR)} incorporates knowledge entities and relations represented by a knowledge graph embedding algorithm (i.e., TransR~\cite{Lin2015LearningEA}).
    
\end{itemize}

\subsubsection{Implementation:}
We implement the above models with PyTorch and Huggingface Transformers\footnote{\url{https://github.com/huggingface/transformers}} libraries.
In Seq2Seq and HRED baselines, we employ GRU architecture~\cite{Cho2014LearningPR}  as the encoder and the decoder with 200 hidden cells.
In terms of word embeddings, we adapt Tencent AI Lab word embeddings of 200d\footnote{\url{https://ai.tencent.com/ailab/nlp/en/embedding.html}}.
When encoding context, all models treat the concatenation of the past $n-1$ utterances as the input of the encoder, while the target output of the decoder is the $n$-th utterance. $n$ is set to 8 in our experiments suggested by KdConv~\cite{Zhou2020KdConvAC}.
All models are optimized with ADAM optimizer using an initial learning rate of 5e-5. 
The mini-batch size is set to 32. 

For our \texttt{RT-KGD}, the embedding size of entities and relations is set to 200. The implementation of TransR is provided by \textit{OpenKE}\footnote{\url{https://github.com/thunlp/OpenKE}}.
The knowledge encoder is Bi-GRU, the hidden size and the number of layers are set to 300 and 1, respectively.
We choose the \textit{Chinese BART}\footnote{\url{https://huggingface.co/fnlp/bart-base-chinese}} as the baseline pre-training language model with the default hyper-parameter settings. 
%
When decoding the response, the beam search size of all models is set to 5. The $\lambda$ is set to 1 in Eq.~\ref{eq:loss}.

\begin{table*}[t]
    \centering
    \resizebox{0.8\textwidth}{!}{
    \begin{tabular}{l|c|cccc|cccc}
    \toprule
    \textbf{Model} & \multicolumn{1}{c|}{\textbf{PPL} $\downarrow$} & \multicolumn{4}{c|}{\textbf{BLEU-1/2/3/4} $\uparrow$} & \multicolumn{4}{c}{\textbf{Distinct-1/2/3/4} $\uparrow$} \\
    \midrule
    \multicolumn{10}{c}{\textbf{Film}} \\
    \midrule
    \textbf{Seq2Seq} & 23.88$^\dagger$ & 26.97$^\dagger$ & 14.31$^\dagger$ & 8.53$^\dagger$ & 5.30$^\dagger$ & 2.32$^\dagger$ &	6.13$^\dagger$ &	10.88$^\dagger$ &	16.14$^\dagger$ \\
    \textbf{Seq2Seq+know} & 25.56$^\dagger$ & 27.45$^\dagger$ &	14.51$^\dagger$ &	8.66$^\dagger$ & 5.32$^\dagger$ & 2.85$^\dagger$ & 7.98$^\dagger$ & 15.09$^\dagger$ & 23.17$^\dagger$ \\
    \textbf{HRED} & 24.74$^\dagger$ &	27.03$^\dagger$ &	14.07$^\dagger$ &	8.30$^\dagger$ &	5.07$^\dagger$ &	2.55$^\dagger$ &	7.35$^\dagger$ &	14.12$^\dagger$ &	21.86$^\dagger$ \\
    \textbf{HRED+know} & 26.27$^\dagger$ & 27.94$^\dagger$ & 14.69$^\dagger$ & 8.73$^\dagger$ &	5.40$^\dagger$ & 2.86$^\dagger$ & 8.08$^\dagger$ & 15.81$^\dagger$ & 24.93$^\dagger$ \\
    \cmidrule{1-10}
    \textbf{BART} & \textbf{2.66} &	28.54 &	19.28 &	14.21 &	11.00 &	2.46 &	14.12 &	25.72 &	36.12 \\
    \textbf{BART+know} & 2.85 &	29.38 &	20.18 &	15.02 &	11.74 &	2.55 &	15.26 &	28.01 &	39.45  \\
    \textbf{BART+know(TransR)} & 2.82 &	29.68 &	20.43 &	15.26 &	11.97 &	2.50 &	15.12 &	27.96 &	39.56 \\
    \cmidrule{1-10}
    \textbf{RT-KGD(ours)} & 2.86 &	\textbf{32.11} & \textbf{22.21} & \textbf{16.68} &	\textbf{13.18} & \textbf{3.05} & \textbf{16.34} & \textbf{31.36} & \textbf{44.68} \\
    \midrule
    \multicolumn{10}{c}{\textbf{Music}}\\
    \midrule
    \textbf{Seq2Seq} & 16.17$^\dagger$ & 28.89$^\dagger$ & 16.56$^\dagger$ & 10.63$^\dagger$ & 7.13$^\dagger$ & 2.52$^\dagger$ & 7.02$^\dagger$ &	12.69$^\dagger$ &	18.78$^\dagger$ \\
    \textbf{Seq2Seq+know} & 17.12$^\dagger$ &	29.6$^\dagger$ & 17.26$^\dagger$ & 11.36$^\dagger$ & 7.84$^\dagger$ & 3.93$^\dagger$ & 12.35$^\dagger$ & 23.01$^\dagger$ & 34.23$^\dagger$ \\
    \textbf{HRED} & 16.82$^\dagger$ &	29.92$^\dagger$ &	17.31$^\dagger$ &	11.17$^\dagger$ &	7.52$^\dagger$ & 2.71$^\dagger$ & 7.71$^\dagger$ & 14.07$^\dagger$ & 20.97$^\dagger$ \\
    \textbf{HRED+know} & 17.69$^\dagger$ & 29.73$^\dagger$ & 17.51$^\dagger$ & 11.59$^\dagger$ & 8.04$^\dagger$ &	3.80$^\dagger$ & 11.70$^\dagger$ & 22.00$^\dagger$ & 33.37$^\dagger$ \\
    \cmidrule{1-10}
    \textbf{BART} & 2.46 & 31.65 & 23.04 & 18.22 & 15.05 & 2.80 & 13.69	& 24.73	& 34.59 \\
    \textbf{BART+know} & \textbf{2.40} &	32.20 &	23.24 &	18.20 &	14.89 &	2.74 &	13.54 &	24.96 &	35.41\\
    \textbf{BART+know(TransR)} & 2.44 &	32.27 &	23.40 &	18.44 &	15.22 &	2.80 &	13.68 &	25.19 &	35.61\\
    \cmidrule{1-10}
    \textbf{RT-KGD(ours)} & 2.47 & \textbf{40.75} & \textbf{31.26} & \textbf{25.56} & \textbf{21.64} &	\textbf{4.18} &	\textbf{17.38} & \textbf{30.05} & \textbf{41.05}\\
    \midrule
    \multicolumn{10}{c}{\textbf{Travel}}\\
    \midrule
    \textbf{Seq2Seq} & 10.44$^\dagger$ &	29.61$^\dagger$ &	20.04$^\dagger$ &	14.91$^\dagger$ &	11.74$^\dagger$ &	3.75$^\dagger$ &	11.15$^\dagger$ &	19.01$^\dagger$ &	27.16$^\dagger$ \\
    \textbf{Seq2Seq+know} & 10.62$^\dagger$ &	37.04$^\dagger$ &	27.28$^\dagger$ &	22.16$^\dagger$ &	18.94$^\dagger$ &\textbf{	4.25$^\dagger$} & 13.64$^\dagger$ & 24.18$^\dagger$ & 34.08$^\dagger$ \\
    \textbf{HRED} & 10.90$^\dagger$ &	30.92$^\dagger$ &	20.97$^\dagger$ &	15.61$^\dagger$ &	12.30$^\dagger$ &	4.15$^\dagger$ & 12.01$^\dagger$ & 20.52$^\dagger$ & 28.74$^\dagger$ \\
    \textbf{HRED+know} & 11.15$^\dagger$ & 36.87$^\dagger$ & 26.68$^\dagger$ & 21.31$^\dagger$ & 17.96$^\dagger$ & 3.98$^\dagger$ & 13.31$^\dagger$ & 24.06$^\dagger$ & 34.35$^\dagger$ \\
    \cmidrule{1-10}
    \textbf{BART} & 1.83 & 34.77 & 29.11 & 25.69 & 23.33 & 2.70 & 13.39 & 21.92 & 29.53 \\
    \textbf{BART+know} & 1.67 &	36.19 &	29.83 &	26.04 &	23.41 &	2.59 &	13.31 &	22.01 &	29.69\\
    \textbf{BART+know(TransR)} & 1.69 &	36.61 &	30.29 &	26.54 &	23.92 &	2.56 &	13.58 &	22.85 &	30.87\\
    \cmidrule{1-10}
    \textbf{RT-KGD(ours)} & \textbf{1.61} &	\textbf{47.56} & \textbf{41.46} & \textbf{37.40} & \textbf{34.31} & 3.58 & \textbf{15.50} & \textbf{26.10} & \textbf{35.72} \\
    \bottomrule
    \end{tabular}%
    }
    \setlength{\abovecaptionskip}{0.5cm}
    \caption{Automatic evaluation results on KdConv Corpus. The \textbf{bold} indicates the best performance. The ``+know'' means the models are enhanced by the knowledge base, and the knowledge words are encoded by word embeddings. $\uparrow$ indicates higher is better. $\downarrow$ indicates lower is better. $^{\dagger}$ denotes the results reported by KdConv~\cite{Zhou2020KdConvAC}. }
    \label{auto}
    \vspace{-1cm}
\end{table*}

\subsection{Evaluation Metrics}

\subsubsection{Automatic Evaluation:}
Following~\cite{Zhou2020KdConvAC}, we adopt perplexity (PPL), BLEU scores~\cite{Papineni2002BleuAM}, and Distinct scores~\cite{Li2016ADO} as automatic metrics. In detail,
PPL is used to evaluate whether the generation result is grammatical and fluent. BLEU-n (n=1, 2, 3, or 4) estimates how many n-grams overlap between generated sentences and ground truth references. Distinct-n (n=1, 2, 3, or 4) evaluates the diversity of generated responses.

\subsubsection{Human Evaluation:}
Considering the complexity of the knowledge-grounded dialogue generation task and the limitation of automatic evaluation, it is necessary to further conduct the human evaluation. 
Following KdConv~\cite{Zhou2020KdConvAC}, The criteria of human evaluation include two aspects:
(1) Fluency evaluates whether the generated responses are reasonable and relevant to the given dialogue context.
(2) Coherence measures how relevant the knowledge contained in the generated responses and the counterpart in the ground truth responses.
We randomly select 100 dialogue contexts from KdConv in three domains, respectively, and then ask five well-educated evaluators to judge the generated responses by different models.
The scoring adopts a 3-point scale.

\begin{figure*}[t]
    \centering
	\begin{subfigure}{\linewidth}
		\centering
		\includegraphics[width=0.9\linewidth]{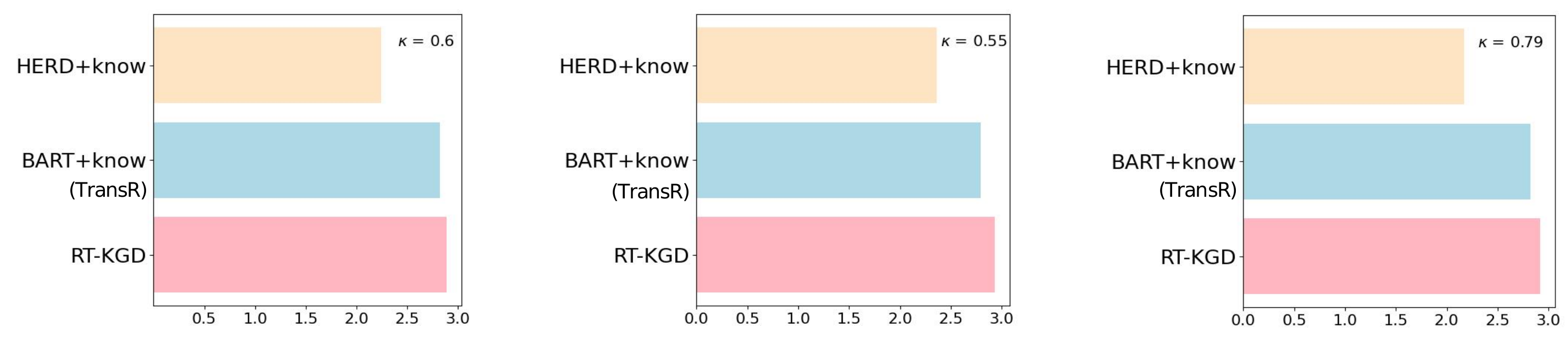}
		\caption{Fluency scores on film, music and travel domains, respectively}
	\end{subfigure}
	
	\begin{subfigure}{\linewidth}
		\centering
		\includegraphics[width=0.9\linewidth]{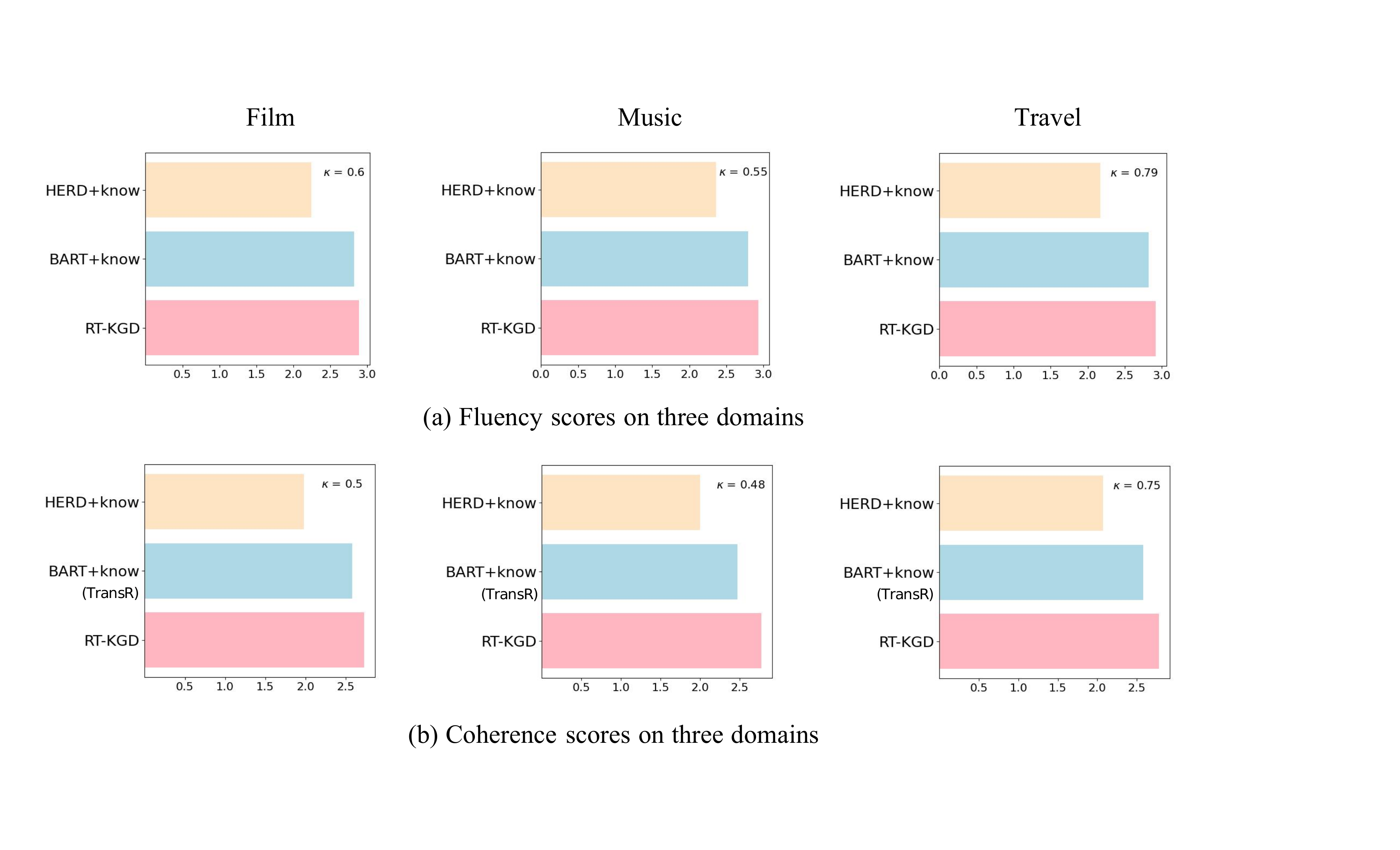}
		\caption{Coherence scores on film, music and travel domains, respectively}
	\end{subfigure}
	\caption{Human evaluation in three domains, including means and variances of the Fluency (a) and Coherence (b). $\bm{\kappa}$ is the Fleiss’ kappa value.}
	\label{fig:human}
\end{figure*}

\subsection{Experimental Results}

Tab.~\ref{auto} shows the automatic evaluation results. We analyze the results from the following perspectives:

\textbf{(1) Comparison between models:}
Compared with all baseline models, \texttt{RT-KGD} achieves the best results on most of automatic metrics in three domains, which indicates that our knowledge-guided method is extremely effective in improving the coherence and diversity of generated responses.
Specifically, compared with Seq2Seq-based and HRED-based models, our \texttt{RT-KGD} obtains not only lower PPL scores but also higher BLEU-n and Distinct-n scores in three domains.
This is because we utilize the pre-trained language model to encode contexts and generate responses, which makes use of the implicitly learned knowledge from the pre-trained corpus.
%
On the other hand, compared with BART-based models, our \texttt{RT-KGD} works better in terms of BLEU-n and Distinct-n scores, however worse on PPL scores.
Based on our manual sampling analysis of the experimental results, the reason might be that our \texttt{MHKT-Path} takes the knowledge transition into consideration.
%
At the same time, diverse knowledge information may result in responses that have never appeared in the corpus, thus reducing the PPL scores.

Moreover, it can be seen that all models with knowledge perform better than those without knowledge in terms of BLEU-n and Distinct-n, indicating the benefits of incorporating knowledge. 
However, the addition of knowledge works worse in PPL. The reason may be that the sentence with knowledge is less common and more difficult to understand for the model. 
We also observe that all models with ``know(TransR)'' obtain higher BLEU-n and Distinct-n scores than models with ``know'', demonstrating that introducing of knowledge graph embedding algorithm has a positive influence on generating high-quality responses.
It is worth noting that in the music domain, BART performs better than BART+know in terms of Distinct-1 and Distinct-2 but worse in Distinct-3 and Distinct-4, which is due to  that BART prefers to use individual words with low frequency rather than common phrases.
Furthermore, it is possible to get a high Distinct-1 for putting together a response with entirely random words.
The same analysis comparing \textit{BART} and \textit{BART+know} also applies to the travel domain.

\vspace{0.5ex}
\textbf{(2) Comparison between domains:}
As we can see, models in the travel domain perform better than that in film and music domains on PPL and BLEU-k, while models in the film domain obtain higher Distinct-n scores than the same model in music and travel domains. The reason might be that there are more entities and relations in the film domain, which leads to more diverse knowledge tokens but a lower similarity with the ground-truth.

\subsection{Human Study}

Here, we estimate three knowledge-grounded dialogue generation models which perform better than other baselines.
The experiment results are shown in Fig.~\ref{fig:human}.
As can be seen, \texttt{RT-KGD} outperforms other models significantly on both metrics in all three domains, which indicates that our model can generate more human-like responses.
Moreover, the performance gap between models behaves differently on different metrics. 
The fluency scores in the music domain (the middle one in Fig.~\ref{fig:human}(a)) are increased from 1.36 (HRED+know) to 1.93 (\texttt{RT-KGD}), while the coherence scores in the music domain (the middle one in Fig.~\ref{fig:human}(b)) are increased from 1.00 (HRED+know) to 1.77 (\texttt{RT-KGD}).
We also show Fleiss' Kappa values of our human study. A higher score indicates higher agreements among evaluators. The kappa scores demonstrate a good inter-agreement among our evaluators.

\subsection{Ablation Study}

\begin{table*}[t]
    \centering
    \resizebox{0.8\textwidth}{!}{
    \begin{tabular}{l|c|cccc|cccc}
    \toprule
    \textbf{Model} & \multicolumn{1}{c|}{\textbf{PPL} $\downarrow$} & \multicolumn{4}{c|}{\textbf{BLEU-1/2/3/4} $\uparrow$} & \multicolumn{4}{c}{\textbf{Distinct-1/2/3/4} $\uparrow$} \\
    \midrule
    \multicolumn{10}{c}{\textbf{Film}} \\
    \midrule
    \textbf{RT-KGD(ours)} & 2.86 &	\textbf{32.11} &	\textbf{22.21} &	\textbf{16.68} &	\textbf{13.18} &	\textbf{3.05} &	\textbf{16.34} &	\textbf{31.36} &	\textbf{44.68} \\
    \quad \textbf{- w/o tri} & \textbf{2.85} &	30.17 &	20.82 &	15.58 &	12.22 &	2.61 &	15.79 &	\underline{29.28} &	\underline{41.16}  \\
    \quad \textbf{- w/o rel} & \underline{3.37} &	\underline{30.10} &	\underline{20.64} &	\underline{15.42} &	\underline{12.10} &	2.56 &	\underline{15.76} &	29.31 &	41.44 \\
    \quad \textbf{- w/o edge} & 3.35 &	30.13 &	20.76 &	15.52 &	12.22 &	\underline{2.53} &	15.79 &	29.42 &	41.68\\
    \midrule
    \multicolumn{10}{c}{\textbf{Music}}\\
    \midrule
    \textbf{RT-KGD(ours)} & 2.47 &	\textbf{40.75} &	\textbf{31.26} &	\textbf{25.56} &	\textbf{21.64} &	\textbf{4.18} &	\textbf{17.38} &	\textbf{30.05} & \textbf{41.05} \\
    \quad \textbf{- w/o tri} & 2.43 &	\underline{32.22} &	\underline{23.24} &	\underline{18.22} &	\underline{14.94} &	\underline{2.74} &	\underline{13.17} &	\underline{24.26} &	\underline{34.42} \\
    \quad \textbf{- w/o rel} & \underline{2.49} & 32.53 & 23.66 & 18.67 & 15.44 & 2.85 &	14.12 &	26.28 &	37.22 \\
    \quad \textbf{- w/o edge} & \textbf{2.42} &	32.28 &	23.44 &	18.50 &	15.26 &	2.83 &	13.92 &	25.36 &	35.55 \\
    \midrule
    \multicolumn{10}{c}{\textbf{Travel}}\\
    \midrule
    \textbf{RT-KGD(ours)} & \textbf{1.61} &	\textbf{47.56} &	\textbf{41.46} &	\textbf{37.40} &	\textbf{34.31} &	\textbf{3.58} &	\textbf{15.50} &	\textbf{26.10} & \textbf{35.72} \\
    \quad \textbf{- w/o tri} & 1.70 &	\underline{36.92} &	30.69 &	26.95 &	24.33 &	2.71 &	13.89 &	23.32 &	31.76 \\
    \quad \textbf{- w/o rel} & \underline{1.84} &	36.98 &	\underline{30.59} &	\underline{26.74} &	\underline{24.06} &	2.64 &	13.63 &	23.01 &	31.17 \\
    \quad \textbf{- w/o edge} & 1.82 &	37.39 &	31.02 &	27.21 &	24.55 &	\underline{2.58} &	\underline{13.43} &	\underline{22.14} &	\underline{29.79} \\
    \bottomrule
    \end{tabular}%
    }
    \setlength{\abovecaptionskip}{0.5cm}
    \caption{Ablation study on KdConv. The \textbf{bold} and \underline{underline} denote the best and the worst performances, respectively.}
    \label{tab:ab}
\end{table*}

To analyze which components are driving the improvements, we further design three graph variants for detailed comparison and ablation study: (1) ``w/o tri'' removes the triplet vertices in \texttt{MHKT-Path}; (2) ``w/o rel'' removes the relation vertices in \texttt{MHKT-Path}; (3) ``w/o edge'' removes the edges between the triplet and the relation vertices in \texttt{MHKT-Path}.

Tab.~\ref{tab:ab} shows the results of ablation studies.
First, we observed that models suffer the performance drop when removing any of the components, demonstrating the effectiveness of integrating triplets and relations.
Second, the degree of impact increases from the film domain to the travel domain after removing components.
For example, the BLEU-n scores decrease by 1.4, 7.4, and 10.4 on average in film, music, and travel, respectively, which shows that our \texttt{MHKT-Path} plays a more significant role in the travel domain in improving the quality of generated response.
Third, the contribution of each component is not equal in different domains.
Specifically, if the triplet vertices are removed, BLEU-n and Distinct-n scores are dramatically dropped in the music domain, indicating that turn-level entity information is capable of enhancing knowledge comprehension.
While removing the relation vertices, BLEU-n scores declined most significantly in film and travel domains, demonstrating the advantage of explicitly modeling dialogue-level relation transition regularities.
Lastly, without the edges between the triplet and relation vertices, the performance of \texttt{RT-KGD} in all three domains is reduced to varying degrees. This is because the edge between triplet vertices and relation vertices effectively propagates the information between these two vertices.

\subsection{Case study}

\begin{figure}[t]
    \centering
    \includegraphics[width=0.95\textwidth]{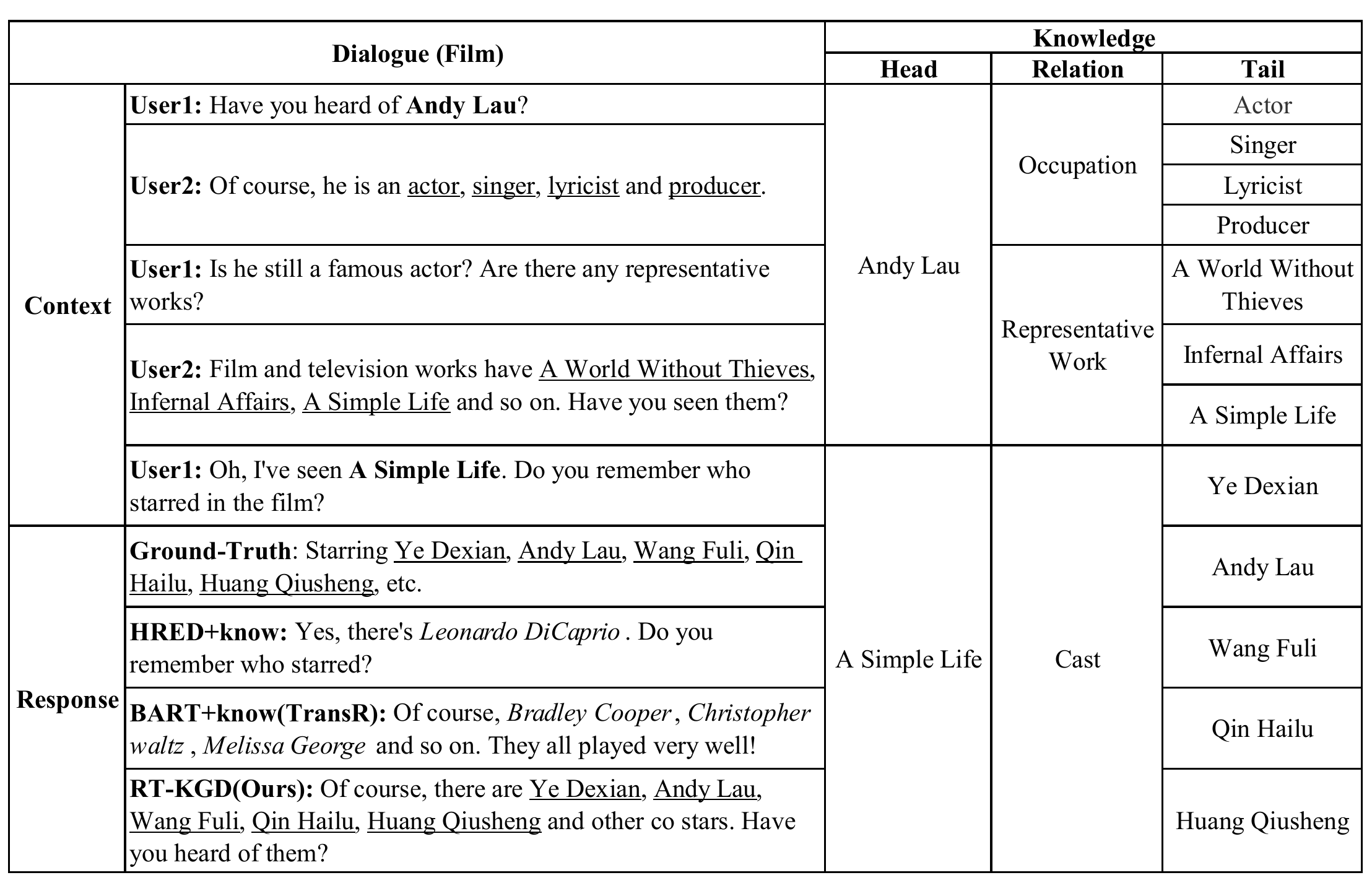}
    \caption{Example dialogue cases. The \textbf{bold} is the core entity under discussion. \underline{Underline} is the appropriate knowledge used in the dialogue. \textit{Italic} is inconsistent with the context.}
    \label{fig:case}
\end{figure}

As shown in Fig.~\ref{fig:case}, we show the responses generated by HRED+know, BART\\+know(TransR) and \texttt{RT-KGD}. We can observe that given the context and corresponding knowledge triplets, HRED+know tends to generate generic or irrelevant responses, and BART+know(TransR) can generate coherent and informative responses but utilizes the inconsistent knowledge. While our \texttt{RT-KGD} is superior to generating high-quality responses with appropriate knowledge.

\section{Conclusion}
\label{sec:5}
In this paper, we proposed a novel KGD model: Relation Transition aware Knowledge-Grounded Dialogue Generation (\texttt{RT-KGD}), which models the knowledge transition across multi-turn dialogue by integrating dialogue-level relation transition regularities with turn-level entity semantic information. Furthermore, our \texttt{RT-KGD} model utilizes the predicted knowledge to generate a response given the dialogue context. According to automatic and manual evaluation, our model generates high-quality responses which utilize more appropriate knowledge and are closer to the responses given by humans.

\paragraph{\textbf{Acknowledgments:}} 
Zhixu Li and Jianfeng Qu are the corresponding authors. 
This research is supported by the National Natural Science Foundation of China (Grant No. 62072323, 62102276),
the Natural Science Foundation of Jiangsu Province (Grant No. BK20191420, BK20210705, BK20211307), 
the Natural Science Foundation of Educational Commission of Jiangsu Province, China (Grant No. 21KJD520005),
the Major Program of the Natural Science Foundation of Jiangsu Higher Education Institutions of China (Grant No. 19KJA610002),\\
NH33714722 Youth Team on Interdisciplinary Research Soochow University - Research on Subjectivity and Reasoning Theory in Artificial Intelligence,
the Priority Academic Program Development of Jiangsu Higher Education Institutions, Suda-Toycloud Data Intelligence Joint Laboratory, and the Collaborative Innovation Center of Novel Software Technology and Industrialization.

\paragraph*{Supplemental Material Statement:} The KdConv dataset and part of the baselines in Sec.~\ref{sec:4} are publicly available from Github\footnote{https://github.com/thu-coai/KdConv}. 
Source codes for \texttt{RT-KGD} model are available at \url{https://github.com/tigerwww-git/RT-KGD}.


\bibliographystyle{Springer/splncs04}
\bibliography{references}

\end{document}